# LLM-R: A Framework for Domain-Adaptive Maintenance Scheme Generation Combining Hierarchical Agents and RAG

Laifa Tao[1,2,3,4], Qixuan Huang[2,3,4], Xianjun Wu[6], Weiwei Zhang[2,3,4], Yunlong Wu[2,3,4], Bin Li[2,3,4], Chen Lu[1,2,3,4], Xingshuo Hai[4,5,*]

[1]Hangzhou International Innovation Institute, Beihang University, China

[2]Institute of Reliability Engineering, Beihang University, Beijing, China

[3]Science & Technology on Reliability & Environmental Engineering Laboratory, Beijing, China

[4]School of Reliability and Systems Engineering, Beihang University, Beijing, China

[5]School of Electrical and Electronic Engineering, Nanyang Technological University

[6]Avic landing gear advanced manufacturing co.,ltd.

*Corresponding author, xingshuo.hai@ntu.edu.sg

E-mails Information for Authors: taolaifa@buaa.edu.cn (L.T.); qxhuang@buaa.edu.cn (Q.H.); 505029583@qq.com (X.W.); zww0216@buaa.edu.cn (W.Z.); wuyunlong@buaa.edu.cn (Y.W.); libin1106@buaa.edu.cn (B.L.); luchen@buaa.edu.cn; xingshuo.hai@ntu.edu.sg (X.H.)

## Abstract

The increasing use of smart devices has emphasized the critical role of maintenance in production activities. Interactive Electronic Technical Manuals (IETMs) are vital tools that support the maintenance of smart equipment. However, traditional IETMs face challenges such as transitioning from Graphical User Interfaces (GUIs) to natural Language User Interfaces (LUIs) and managing complex logical relationships. Additionally, they must meet the current demands for higher intelligence. This paper proposes a Maintenance Scheme Generation Method based on Large Language Models (LLM-R). The proposed method includes several key innovations: We propose the Low Rank Adaptation-Knowledge Retention (LORA-KR) loss technology to proportionally adjust mixed maintenance data for fine-tuning the LLM. This method prevents knowledge conflicts caused by mixed data, improving the model's adaptability and reasoning ability in specific maintenance domains. Besides, Hierarchical Task-Based Agent and Instruction-level Retrieval-Augmented Generation (RAG) technologies are adopted to optimize the generation steps and mitigate the phenomenon of hallucination caused by the model's Inability to access contextual information. This enhancement improves the model's flexibility and accuracy in handling known or unknown maintenance objects and maintenance scheme scenarios. To validate the proposed method's effectiveness in maintenance tasks, a maintenance scheme dataset was constructed using objects from different fields. The experimental results show that the accuracy of the maintenance schemes generated by the proposed method reached 91.59%, indicating which improvement enhances the intelligence of maintenance schemes and introduces novel technical approaches for equipment maintenance.

# 1. Introduction

The advancement of intelligent devices has led to increased complexity in equipment and working conditions, making maintenance tasks more demanding for frontline personnel. Consequently, there is a growing need for effective electronic maintenance schemes in production activities[1]. Traditional Interactive Electronic Technical Manuals (IETMs) are proving to be inefficient and outdated. For example, frontline maintenance personnel often generate valuable experiential information during work[2]. However, this data is typically overloaded with redundancy and is slow to update, limiting its effectiveness. Additionally, precise retrieval in IETMs for tasks that involve small samples requires a high level of expertise. It makes the current systems cumbersome and inefficient for managing and customizing maintenance information[3]. In the context of Industry 4.0, the emergence of generative artificial intelligence has significantly changed the way precise searches are conducted within IETMs. This new technology provides stronger support for modern production practices, enhancing the efficiency and effectiveness of maintenance operations[4].

In recent years, extensive research has been conducted on traditional IETMs. For instance, Niu et al.[3] proposed an intelligent maintenance system that incorporates diagnostic strategies such as manual interpretation, rule-based fuzzy reasoning, and condition-based data fusion. Wu et al.[5] proposed a new information query method for IETMs using a Bayesian network model. This model leverages word and unit statistics from the IETM dataset, along with topological model structures and conditional probability learning. It aims to estimate the conditions needed for structured IETM information queries through probabilistic inference. These studies primarily focus on enhancing the efficiency of knowledge and data retrieval in IETMs through innovative technical approaches. However, the increasing complexity of equipment and the demand for efficient information processing presents new challenges. Traditional IETM systems contain extensive textual information describing equipment maintenance operations, which is stored in data modules with specific tags. Users can access and retrieve this textual information through operations on the data modules[6]. However, these systems fail to adequately explore the intrinsic logical relationships within the text, often resulting in incomplete maintenance schemes that overlook current working conditions. Therefore, the interactivity of IETMs is restricted, relying significantly on the thoroughness of technical documentation and the knowledge of technical personnel. This limitation makes it challenging to meet the growing need for intelligent maintenance schemes in complex and demanding environments.

In recent years, Large Language Models (LLMs) have demonstrated exceptional generative and emergent capabilities in the field of artificial intelligence, revolutionizing traditional search paradigms and leading the transition from Graphical User Interfaces (GUIs) to Language User Interfaces (LUIs) in human-computer interaction[7]. By collecting data from specific domains and applying fine-tuning techniques[8], LLMs can deeply understand domain-specific information and align more closely with human preferences. For instance, Huang et al.[9] incorporated legal knowledge during the continued pre-training stage and used supervised fine-tuning tasks to create a legal domain LLM. Wen et al.[10] combined general and domain-specific datasets for fine-tuning, enhancing the model's expertise in specific areas while maintaining its versatility. Yuan et al.[11] combined a large visual model with the fuzzy DEMATEL method to construct a medical image dataset and a large-scale rehabilitation management model, achieving good results on clinical datasets. Retrieval-Augmented Generation (RAG) technology[12] has been widely discussed by researchers in the process of continued iteration of LLM technologies. Endowed with the retrieval and generation capabilities of LLMs, RAG can provide external database information to enhance the model accuracy and information richness when encountered with specific domains. Eric Melz[13] proposed the ARM-RAG method, which improves LLM performance without incurring high training costs. Li et al.[14] integrated RAG with upstream datasets and downstream performance evaluation, enhancing the factual accuracy of LLMs for specific domain and time-sensitive queries related to private knowledge bases. Furthermore, LLMs possess powerful capabilities in knowledge acquisition, instruction comprehension, generalization, and reasoning but have limitations in external execution abilities. LLM capabilities can be further enhanced by Agents. Agents not only extends the perception and action space of LLMs, but also broadens human-machine integration through LLM reasoning and planning, resulting in deeper human-machine collaboration[15][16]. Liu et al.[17] developed an LLM-Agent framework called DyLAN, designed for collaboration on complex tasks such as reasoning and code generation. Yashar Talebirad et al.[18] modeled various fields, such as court simulations and software development scenarios, through the collaborative work of multiple Agent components with unique attributes and roles, providing a pathway to improve LLM performance. Currently, LLMs and related technologies are under research and application across various fields. They have innovated traditional research and application methods, fundamentally evolving many conventional technical fields as well as human production practices.

Under this background, this paper proposes a Maintenance Scheme Generation Method based on Large Language Models (LLM-R). The method is aligned with the maintenance domain by employing a Low Rank Adaptation-Knowledge Retention (LORA-KR) loss technology. Subsequently, a novel approach combining Hierarchical Task-Based Agent and Instruction-level RAG is proposed to generate maintenance schemes for complex and situations with small sample data. This method addresses the limitations of traditional search paradigms and maintenance schemes, enabling intelligent assisted maintenance.

The main innovations of this paper are as follows:

(1) This study proposes a Supervised Fine-Tuning of LORA-KR loss, which leverages proportionally mixed data in an innovative way to enhance the model's performance in the maintenance domain. In this method, we fine-tune the model by combining domain-specific datasets with general knowledge datasets using the optimal ratio calculated by the KR loss method. This approach mitigates catastrophic forgetting while preserving the model's original reasoning capabilities. As a result, the model generates more accurate and practical maintenance information, reducing dependence on biased or small data sources and lowering the risk of generating false information.

(2) Based on the aforementioned process, this study proposes a maintenance schemes generation framework that combines Hierarchical Task-Based Agents with Instruction-level RAG technology. In this framework, a LLM acts as the central processor, organizing and summarizing information, while various agents supply specific external knowledge as required. This setup allows the framework to decompose complex maintenance tasks and access relevant contextual information more effectively, improving the LLM's ability to handle complex maintenance scenarios and small sample maintenance objects.

(3) This study uses the advanced capabilities of the LLM to overcome the limitations of traditional IETM searches, allowing for conversational generation of maintenance schemes. A generative question-answering system was developed to accurately retrieve and provide schemes highly relevant to maintenance tasks. In particular, even in complex environments with small sample data, this system effectively offers tips and guidance for related tasks. Therefore, it enhances both the accuracy and convenience of operations.

The rest of the paper is organized as follows: Section 2 discusses the current research and application of LLM and related technologies. Section 3 presents the LLM-R modeling process, including its definition and six main steps. Section 4 introduces three key methods of LLM-R, including Supervised Fine-Tuning of LORA-KR loss, Hierarchical Task-Based Agent, and Instruction level RAG technology. Section 5 presents the experimental data, model parameters, and baselines. Section 6 presents the experimental results based on maintenance object data, along with the validation and evaluation of the proposed method. Finally, Section 7 summarizes the conclusions and future work.

## 2. Related work

This section first reviews the actual requirements of IETM based on LLMs and identifies the technical weaknesses of current maintenance scheme generation frameworks. Secondly, it summarizes the data mixing strategy and Agent-enhanced RAG technology, discussing their technical feasibility and applicability in various fields. Finally, the gap analysis is carried out.

## 2.1 IETM based on LLMs

In the field of machine learning, LLMs have gained significant attention due to their outstanding performance and vast application potential. Models like GPT-3[19], InstructGPT[20], ChatGLM[21], and Baichuan[22] possess billions or even trillions of parameters. They are capable of capturing deep data features and relationships through extensive datasets and complex network structures. These models demonstrate strong generalization across various tasks and conditions. The self-attention mechanism in the Transformer architecture, which underpins these models, allows for parallel data processing, enhancing computational efficiency and addressing the limitations of traditional neural networks in handling long sequences of data. Combined with these advantages, this paper proposes leveraging the advanced processing capabilities of LLMs to enhance IETM capabilities. The goal is to significantly improve their performance in knowledge management, dynamic updating, accurate interpretation, and standardized generation, thereby solving complex maintenance task generation problems more effectively.

To achieve this, the first priority is to enhance knowledge accuracy and reduce the incidence of hallucinations: IETMs contain extensive numerical information, such as model numbers and dates. However, LLMs generate text based on predicting the probability distribution of the next word. Such a mechanism can lead to numerical confusion when dealing with complex data due to insufficient contextual information or prior knowledge[23]. And then, the performance of LLMs in processing long texts is critical. When interpreting lengthy texts with complex dependencies, models may produce contradictory content[24]. In addition, LLMs often lack sufficient logical reasoning capabilities, which can result in misinterpretations of the source text and incorrect conclusions. For example, models may over-rely on pre-trained knowledge and fail to consider the actual context, leading to conflicts between the context and the built-in knowledge. If the context contains errors or is based on faulty assumptions, LLMs may not recognize these issues and could generate incorrect recommendations. To improve hallucinations in LLMs, various effective methods have been proposed by academia and industry. For example, the use of external knowledge verification to proactively detect and alleviate hallucinations can prevent the spread of hallucinations[25]. Another method involves modifying decoding strategies, such as employing fact-based core sampling algorithms, to dynamically adjust the probability of generated sentences, thereby improving factual accuracy[26]. Additionally, sampling multiple outputs and checking their consistency can help distinguish between factual claims and hallucinations[27].

Secondly, improving domain adaptability and standardization involves addressing specific issues such as dynamic knowledge management, context preservation, concept understanding, and standard adaptability. Detecting knowledge conflicts is challenging because new information can contradict existing content, and LLMs often struggle to detect and resolve these conflicts effectively. The storage and updating of old and new knowledge require a dynamic mechanism, as frequent updates can cause LLMs to lose critical context. When dealing with unknown concepts and terminology, particularly in specialized areas, LLMs can generate inaccurate interpretations.

Additionally, IETMs need to adhere to industry-specific standards to ensure content consistency and compatibility, posing a significant challenge for LLMs to adapt to different standards and generate compliant content. To solve these problems, research has focused on dynamic knowledge management systems and the construction of external knowledge channels. For instance, the knowledge conflict detection capability of LLMs is enhanced by constructing domain-specific knowledge graphs and trees[28]. RAG model is employed to bridge the information gap in LLMs, improving the understanding and application of domain-specific knowledge[29]. In addition, using external knowledge bases or memory retrieval mechanisms ensures the accuracy and consistency of the generated content[30].

Despite significant advancements in these technologies, limitations still constrain the development of LLM-based IETMs. For example, improving knowledge accuracy is highly dependent on data quality, and low-quality data can directly affect model performance[31]. In addition, tasks such as knowledge conflict detection, joint training, and retrieval are complex. Consequently, it is difficult to solve these problems end-to-end, requiring a large amount of high-quality data and fine pipe design. This study addresses these problems by summarizing the data sample mixing strategy and Agent-enhanced RAG technology from the perspectives of improving data quality and task pipeline design. It discusses the technical feasibility and field applicability of the proposed Supervised Fine-Tuning of LORA-KR loss, Hierarchical Task-Based Agents, and Instruction-level RAG technology within the LLM-R framework.

## 2.2 Data sample mixing strategy

The data mixing strategy has demonstrated technical feasibility in expanding sample knowledge coverage, improving pattern recognition accuracy, and enhancing LLM efficiency. This approach makes models more robust with stronger generalization capabilities and has been widely applied to various tasks[32], including image, time series, and language generation, aiming to produce high-quality training data. Data sample mixing strategies can be categorized into two main types[33]: one is partial erasure, mixing, patch, label mixing, etc., based on the original data; the other is mixing on the sampling of the original data. The mixing strategy based on raw data includes the following methods: erasing class methods, such as hiding part of the original sample to enhance the learning of contextual information[34]; The pixel-based mixing method, for example, creates a new training sample by weighted averaging the pixels of the two images, linearly combines the associated labels appropriately by linear interpolation of the feature vectors, and builds a mixer to parameterize the mixing strategy[35]. Feature space mixing enhances data through the hidden layer feature representation of mixed samples[36]. Sampling-based strategies involve methods such as weighted mixed sampling, which assigns importance-based weights to different data sources to optimize mixing and improve model performance prediction[37]. Strategic sampling selects data from various datasets in a way that oversamples small but important datasets or under samples large but less relevant ones. Transfer learning and domain adaptation technologies enable generic models to better fit domain-specific data by leveraging existing model knowledge to adjust the effects of different data sets. Dynamic data enhancement expands the dataset by adjusting the enhancement strategy artificially based on the model's current performance, especially when specific domain data is limited[38].

## 2.3 Agent-enhanced RAG technology

Traditional LLM interaction models require users to provide detailed instructions and background information, which can be limiting, especially when users do not fully understand the requirements of complex tasks. To overcome this challenge, the concept of intelligent agents has been introduced. These agents can understand and execute more complex instructions than direct commands, providing users with a more efficient and dynamic interactive experience. The architecture of an agent is divided into three parts: brain (cognitive processing unit), perception (receiving unit of environmental input), and action (executive unit). Based on the Langchain framework[39], researchers have designed agents with various functions to better adapt to specific field requirements. For example, Wu et al. promoted autonomous cooperation among agents through role-playing methods[40]. Kannan et al. proposed the SMART-LLM framework to transform high-level task instructions into multi-robot task planning[41]. Zhou et al proposed a new framework of TRAD[42], which builds an LLM-based agent using a track-level retrieval method to improve performance in sequential decision-making tasks. At the same time, RAG technology, which integrates external databases with LLMs, is particularly suitable for knowledge-intensive applications. Melz et al. proposed ARM-RAG to address knowledge forgetting and inference memory limitations by combining RAG with LLMs[43]. Wang et al. achieved new best results on the Common Gen test benchmark by enhancing common knowledge generation tasks with a new search-based framework[44]. RAG technology works by converting data from various formats into a standardized format during the indexing phase and processing it vectorially through model embeddings. These text blocks and their vector embeddings are stored as key-value pairs, forming a scalable and easily retrievable maintenance scheme vector library. In the retrieval phase, RAG uses a BERT encoder for maximum inner product search (MIPS) to select relevant text blocks for generation. The generator then combines query questions with selected document sets to produce output. By integrating retrieval and generation methods, this approach significantly improves the accuracy and reliability of data, showcasing great potential and practical value in knowledge-intensive fields. Combining agent-enhanced RAG technology not only improves user interaction but also enhances the efficiency of solving complex tasks.

## 2.4 Gap analysis

The data mixing strategy and Agent proposed in this paper demonstrate significant technical feasibility. By optimizing the weight allocation of different data sources, the data mixing strategy enhances the model's adaptability and generalization capabilities in complex and dynamic environments. Meanwhile, the Agent architecture allows for more precise task control and effective problem-solving strategies, thereby improving data quality and task pipeline design. This combination advances the development of LLM-based IETM.

## 3. The process of generating maintenance schemes

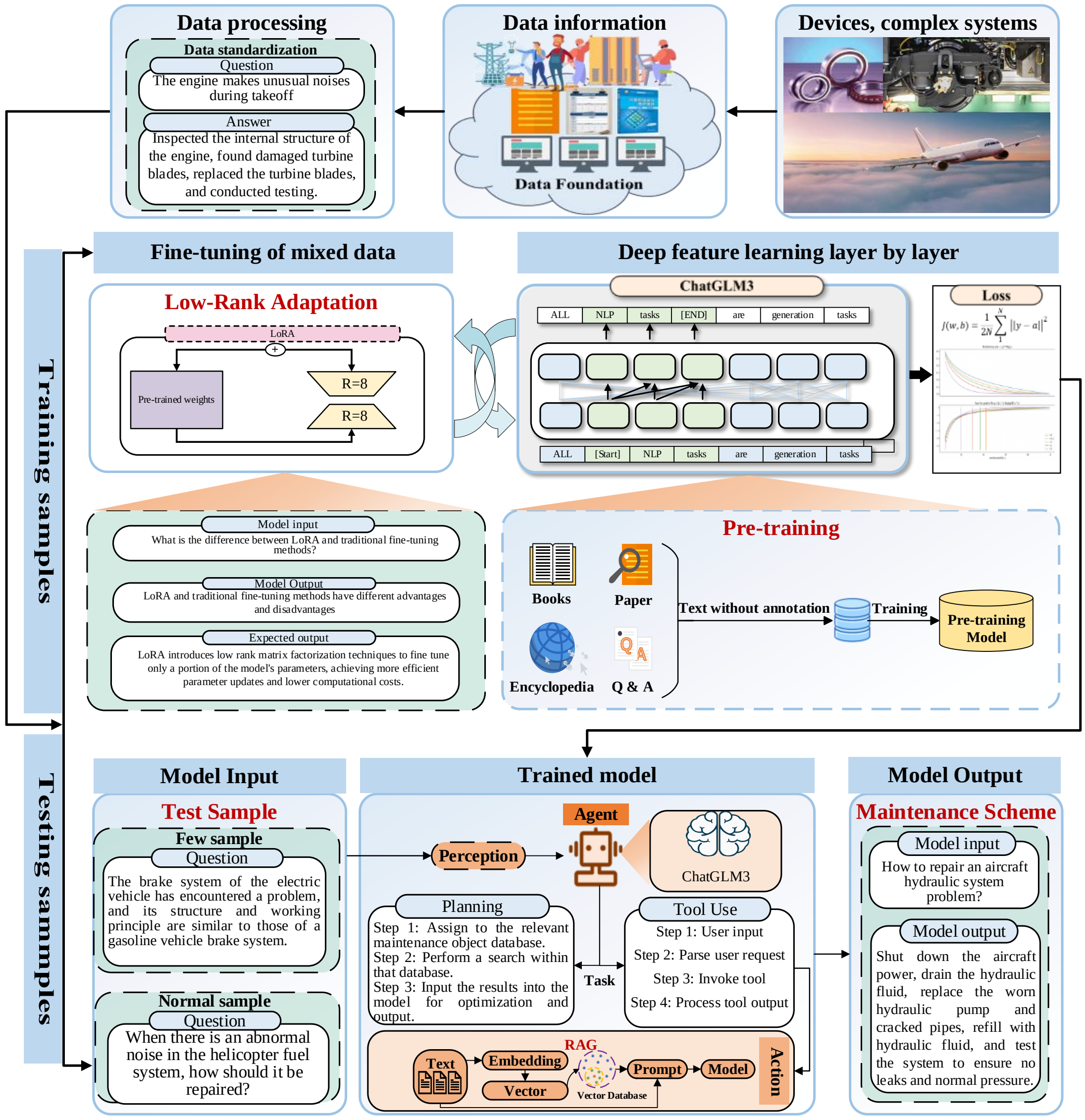

Figure 1. The framework of the proposed method.

In this paper, LLM-R is used to analyze complex maintenance tasks in a highly customized manner, and the corresponding maintenance schemes can be accurately retrieved. This approach optimizes the mechanical input issues present in traditional IETMs, significantly improving the efficiency of maintenance personnel in handling complex tasks. Unlike traditional IETMs, where maintenance schemes are often fixed and cannot adapt to current needs, LLM-R analyzes both the maintenance object and the scheme to provide adaptable solutions. In particular, when faced with complex maintenance tasks and unknown maintenance objects, LLM-R employs Hierarchical Task-Based Agents and Instruction-level RAG technology to analyze the tasks and retrieve maintenance schemes for similar objects based on the maintenance base LLM. This demonstrates the model's ability to learn from small samples, showing that LLM-R can effectively handle situations that traditional IETMs cannot. This dynamic optimization ensures that the maintenance schemes are aligned with the actual conditions and needs of the maintenance process, thereby improving maintenance efficiency. The process for generating maintenance schemes using the LLMs proposed in this paper is shown in Figure 1, and the specific process is as follows:

(1) Extraction of maintenance data: The first step in the proposed method is to extract relevant maintenance data from the equipment's maintenance process. This data includes detailed historical maintenance records, maintenance procedure documentation, and other relevant technical information. Collecting comprehensive data ensures a rich and accurate dataset for model training and optimization.

(2) Organization of data format: The collected maintenance data needs to be systematically organized to meet the data input format requirements of the LLM. This involves cleaning the data and converting it into a standardized, machine-readable format.

(3) Division of training and test sets: The collated data is divided into training and test sets. The training set is used to train the model, while the test set evaluates the model's performance. Ensuring that the data is representative during partitioning is crucial for the model to fully learn the semantic and logical relationships between various maintenance objects and schemes.

(4) Base model: This step involves choosing a base LLM for further training and fine-tuning. The selection should account for the specific needs and data characteristics of the maintenance field. Factors to be considered include the language support, coverage, and architecture of the pre-trained dataset. The chosen LLM must effectively support the generation of maintenance schemes and possess sufficient scalability to adapt to various maintenance tasks across different objects.

(5) Training model: Once the LLM is selected, the pre-trained Transformer model is fine-tuned using Supervised Fine-Tuning of LORA-KR loss, combined with Hierarchical Task-Based Agents and Instruction-level RAG technology. LoRA optimizes the low-rank adaptation layer, adjusting part of the weight by minimizing the loss function to align with the special domain. The Hierarchical Task-Based Agent uses the LLM as the "brain" to decompose maintenance tasks and employs Instruction-level RAG technology to incorporate external information.

(6) Test and evaluation: The model is tested using the test data, with a comprehensive evaluation conducted using formulas, language models, and manual assessments to ensure applicability in real maintenance scenarios. The test data includes known and unknown maintenance objects/scenarios to fully assess LLM-R's generalization capability under normal and small data inputs.

## 4. Method

This section explores the structure and components of the LLM-R, as shown in Figure 2. LLM-R has a basic LLM that is fine-tuned through Supervised Fine-Tuning of LORA-KR loss, while LLM-R includes Hierarchical Task-Based Agents and Instruction-level RAG technology.

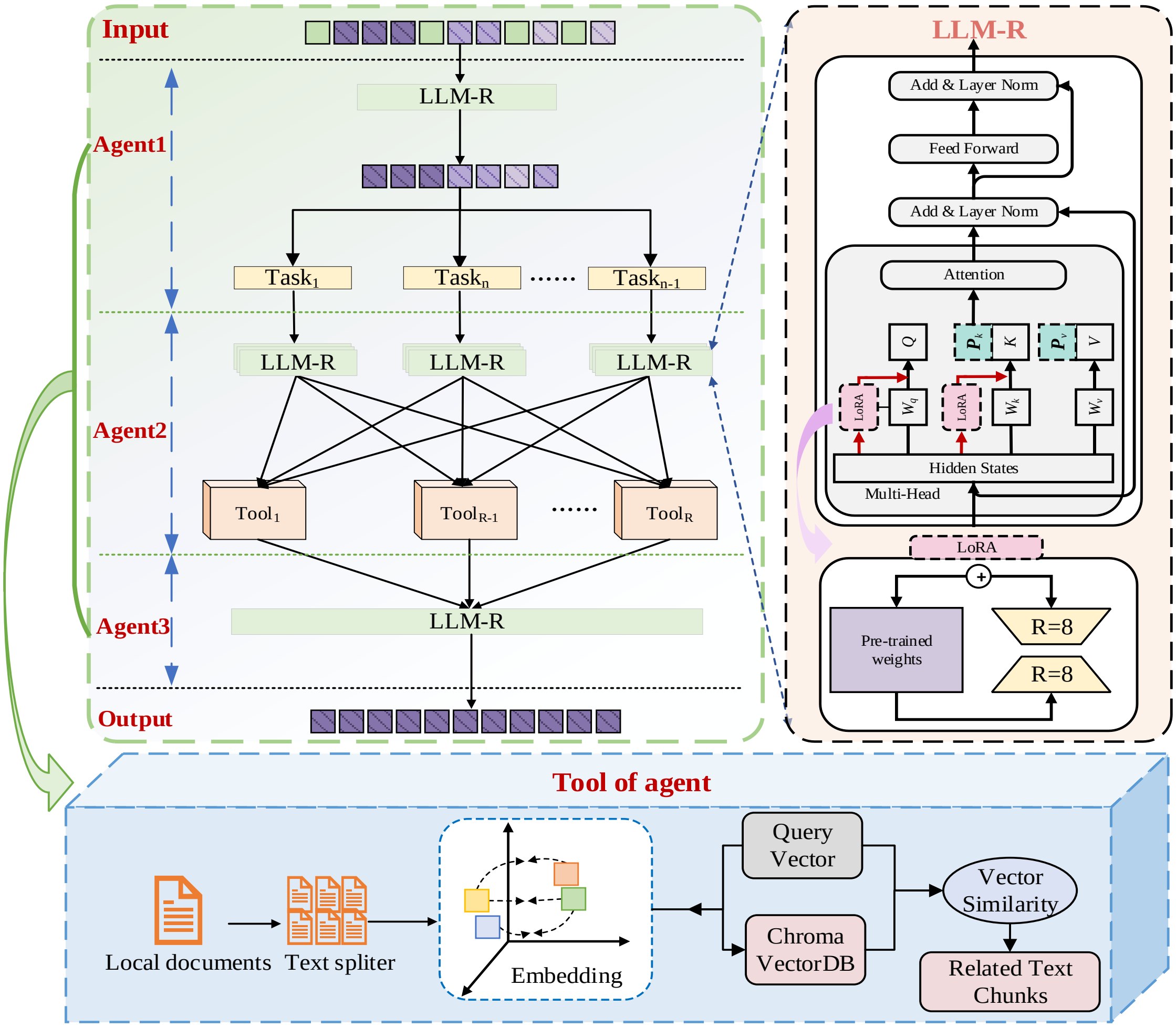

Figure 2. The framework of LLM-R.

A base LLM specifically designed for maintenance tasks was created through fine-tuning of the ChatGLM3-6B large language model. Utilizing a Supervised Fine-Tuning of LORA-KR loss technology, the model transfers its generative capabilities to vertical domain without compromising its original abilities. The fine-tuned base LLM plays a "scheduling and optimization" core role in the LLM-R system. By Hierarchical Task-Based Agent technology and Instruction-level RAG technology, it parses user requirements, decomposes maintenance tasks into subtasks, and assigns them to the appropriate tools for processing. By integrating Agent and RAG technologies, the model not only improves the accuracy of maintenance schemes generation but also demonstrates small samples learning capability, enabling it to rapidly generate effective solutions even for new maintenance tasks.

The LLM-R framework proposed in this study consists of three key components. First, the Instruction-level RAG mechanism introduces a retriever (as discussed in Section 4.1) to address novel or unknown maintenance demands by swiftly retrieving the most relevant information and solutions from vast datasets. Second, this paper proposes Supervised Fine-Tuning of LORA-KR loss (as discussed in Section 4.2) to address potential knowledge conflicts that may arise during the integration of domain-specific and general knowledge. This approach prevents knowledge retention conflicts caused by excessive updating of public knowledge when fine-tuning with different proportions of datasets, thus achieving optimal balance between domain-specific and general knowledge. Finally, during the model execution phase, a multi-task Agent module (as discussed in Section 4.3) is designed to precisely analyze user maintenance needs, decompose tasks into multiple subtasks, and assign them to the appropriate tools for efficient processing.

## 4.1 Instruction-level RAG

In the LLM, the effective operation of the Agent relies not only on the powerful processing and organizational capabilities of the LLM itself as the "central brain" but also on using specific tools to execute the Agent's instructions. To enhance the efficiency and accuracy of retrieving specific maintenance object schemes, this study proposes instruction-level RAG technology as the core tool for the Agent.

The specific implementation process of RAG technology is as follows:

After the user or Agent identifies the keywords for the maintenance task, these keywords are input into the matching RAG tool. The instruction-level RAG tool vectorizes these service requirement keywords, converting them into a format that the model can process. These requirement vectors are then matched with vectorized data stored in the database to find the most similar items. Upon completing the matching process, the maintenance scheme or related information that best matches the input requirements is output.

In the retrieval phase, RAG adopts the following retrieval model:

$$p(z|x) \propto \exp\left(BERT_{text}(z)^{\mathrm{T}} BERT_{question}(x)\right)$$

where x and z represent the query vector and the index document, respectively. BERT is used as the encoder, and the first K text blocks with the highest prior probability $p(z|x)$ are retrieved using maximum inner product search (MIPS) to form the set $\mathrm{top}_k(p(\cdot|x))$. Finally, in the generation phase, the generator, represented as $p(y|z,x)$, combines the input of the proposed query question and the selected set of documents to generate the output y. The selected documents are treated as potential variables and marginalized to get the final output distribution:

$$p(y|x) = \sum_{z \in \mathrm{top}_k(p(\cdot|x))} p(y|z,x)\, p(z|x)$$

With the introduction of instruction-level RAG technology, agents can quickly find the most relevant information and solutions from a large amount of data when faced with new or unknown maintenance needs, facilitating the rapid generation and optimization of maintenance solutions. This method significantly enhances the flexibility and efficiency of LLM and Agent applications in the maintenance field, allowing them to more accurately respond to complex and evolving maintenance needs. By combining retrieval and generation, this approach greatly improves the accuracy and reliability of data generation, demonstrating significant potential and practical value in knowledge-intensive applications.

## 4.2 Supervised Fine-Tuning of LORA-KR loss

### 4.2.1 Fine-Tuning

During the fine-tuning process of the LLM-R model, the preparation of the dataset is crucial. To accurately align the LLM with the requirements of the maintenance domain, an innovative data processing method was employed. Two datasets were defined $D_r$, a dataset consisting of question-answer pairs specific to maintenance domain expertise, and $D_c$, a dataset comprising general knowledge question-answer pairs. These two datasets were then integrated and concatenated to form a new dataset, $D_m = W_r D_r \cup W_c D_c$, where $W_r$ and $W_c$ are learnable weights representing the proportions of the data mixture, determined through experimentation as detailed in Section 4.2.2. The retrieval model was fine-tuned using $D_m$ resulting in the output model $M_{finetuned}$. The detailed fine-tuning process is as follows:

According to the problem definition summarized in Section 4.1, we use a BERT encoder to process the query Q, obtaining its vector representation $q = BERT_Q(Q)$. Similarly, the BERT encoder is applied to the entity words E in $D_m$, producing their vector representation $\mathbf{e} = BERT_E(E)$. The query vector q and the entity vector e are then concatenated (or combined using other operations) to form a combined vector $z = \text{Concat}(q, e)$. This combined vector z is passed through a linear classifier to produce the class prediction, $\hat{C} = \text{Softmax}(W_0 \cdot z + b_c)$, where $W_0$ is the weight matrix and $b_c$ is the bias term.

The purpose of LoRA is to fine-tune the weights using low-rank matrices while keeping the main structure of the model intact, thereby reducing computational complexity and memory consumption. During the proportional supervised fine-tuning of the model with mixed data, the updated weight matrix is as follows:

$$W_{0,new} = W_0 + \Delta W = W_0 + BA$$

where $W_0 \in R^{d \times k}$ is the initial weight of ChatGLM3. For LORA, $\Delta W$ fine-tuning is applicable. By dividing $\Delta W$ into two matrices $B \in R^{d \times r}$ and $A \in R^{r \times k}$:

$$\Delta W = B \cdot A$$

the rank $R$ of the matrix is set to 8 when fine tuning. This step greatly reduces the computational load during model training and transfers the capabilities of the original model to the maintenance field.

By substituting the updated weight matrix $W_{0,new}$ into the classifier formula, the new class prediction is obtained as follows:

$$\hat{C} = \text{Softmax}\left((W_0 + B \cdot A) \cdot z + b_c\right)$$

Expanding this, the new class prediction can be expressed as:

$$\hat{C} = \text{Softmax}\,(\mathrm{W}_0 \cdot \mathrm{z} + \mathrm{B} \cdot (\mathrm{A} \cdot \mathrm{z}) + \mathrm{b}_c)$$

Where the second term, $\mathrm{B} \cdot (\mathrm{A} \cdot \mathrm{z})$ represents the low-rank fine-tuning component of the original model's weights introduced by LoRA.

4.2.2 LORA-KR loss

This paper introduces LORA-KR loss, a novel regularization method that integrates domain-specific loss with domain-independent knowledge regularization. The aim is to fine-tune the model without deviating from the pretrained knowledge domain. LORA-KR loss consists of two components: the task-specific loss $\mathcal{L}_{\mathrm{CE}}$ and the domain-independent regularization loss $\mathcal{L}_{\mathrm{KL}}$.

First, the task-specific loss $\mathcal{L}_{\mathrm{CE}}$ is typically a cross-entropy loss, used to adjust the model to fit the data distribution of the specific task.

$$\mathcal{L}_{\mathrm{CE}}(\theta) = -\sum_i y_i \log f(x_i;\theta)$$

Here, $y_i$ represents the true label of the i th sample, and $f(x_i;\theta)$is the model output.

Secondly, the domain-independent regularization loss $\mathcal{L}_{\mathrm{KL}}$ is computed using the KL divergence between the model's output and the pretrained model's output, ensuring that the model retains task-independent knowledge.

$$\mathcal{L}_{\mathrm{KL}}(\theta) = \sum_i y_{\mathrm{pre}}^i \log \frac{y_{\mathrm{pre}}^i}{f(x_i;\theta)}$$

Here, $y_{\mathrm{pre}}^i$ denotes the output of the pretrained model. LORA-KR loss combines the aforementioned two types of losses during LoRA gradient updates to maintain the model's memory of the original pretrained domain knowledge while adapting it to the new task domain.

The formula for LORA-KR loss is as follows:

$$\mathcal{L}_{\text{LORA-KR}} = (1-w) \cdot \mathcal{L}_{\mathrm{CE}} + w \cdot \mathcal{L}_{\mathrm{KL}}$$

Here, $w$ is a weight dynamically adjusted based on the gradient magnitude of the specific sample, used to balance the task-specific loss and the domain-independent regularization loss. The weight $w$ for each sample is dynamically adjusted to balance the contributions of the two losses.

The adjustment formula is as follows:

$$w = \frac{f_t}{f_t + \gamma y_{\mathrm{pre}}^2 / y_t^2}$$

Here, $f_t$is the gradient of the task-specific loss; $y_{\mathrm{pre}}^2$ is the output of the pretrained model, $y_t^2$ is the output of the current model, and $\gamma$ is a hyperparameter used to balance the influence of task-specific knowledge and domain-independent knowledge.

The final LORA-KR loss is optimized by combining the task-specific cross-entropy loss $\mathcal{L}_{\mathrm{CE}}$ and the domain-independent regularization loss $\mathcal{L}_{\mathrm{KL}}$. The formula for LORA-KR loss is as follows:

$$\mathcal{L}_{\text{LORA-KR}} = (1-w) \cdot \mathcal{L}_{\mathrm{CE}} + w \cdot \mathcal{L}_{\mathrm{KL}}$$

Among them: $\mathcal{L}_{\mathrm{CE}}$ is the task-specific cross-pick loss, calculating the difference between the model's predicted class and the true class:

$$\mathcal{L}_{\mathrm{CE}}(\theta) = -\sum_{i=1}^{N} y_i \log \hat{y}_i$$

Here, $y_i$ is the true class label for the ith sample, and $\hat{y}_i$ is the class probability distribution predicted by the model.

$\mathcal{L}_{\mathrm{KL}}$ is a task-independent regularization loss, using KL divergence to measure the difference between the current model output and the pre-trained model output:

$$\mathcal{L}_{\mathrm{KL}}(\theta) = \sum_{i=1}^{N} y_{\mathrm{pre}}^{i} \log \frac{y_{\mathrm{pre}}^{i}}{\hat{y}_i}$$

Here, $y_{\mathrm{pre}}^{i}$ is the predicted probability distribution of the pre-trained model for the i th sample, and $\hat{y}_i$ is the predicted probability distribution of the current model for the i th sample.

### 4.2.3 Optimization of LORA-KR loss

In the retrieval task, we combine LORA-KR loss with LoRA's low-rank matrix update to optimize and regularize the fine-tuning process of the model. This process continues to iterate throughout the training process until the model's loss function converges or a predetermined training cycle is reached. The steps are as Figure 3.

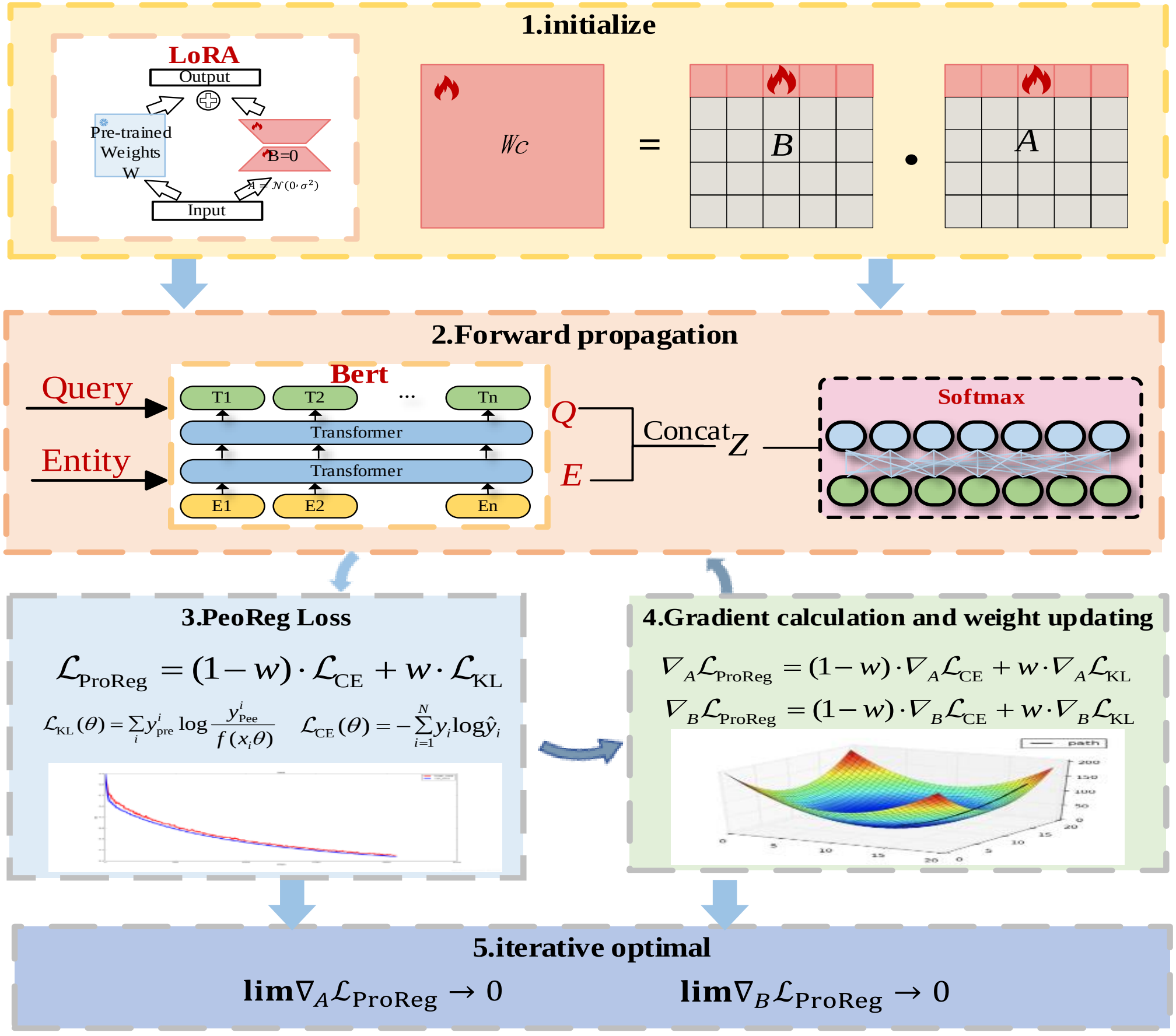

Figure 3. Iterative optimization process

1.Initialization: Initialize LoRA's low-rank matrices $A$ and $B$, and the weight matrix $Wc$.

2.Forward propagation: Enter the query requirement $Q$ and entity word $E$, get vectors representing $q$ and $e$, and then get the category prediction $\hat{C}$.

3.Calculate loss: Calculate LORA-KR loss, including task-specific cross entropy loss $\mathcal{L}_{\mathrm{CE}}$ and task-independent KL divergence regularization loss $\mathcal{L}_{\mathrm{KL}}$.

4.Gradient calculation and weight update: According to the formula of LORA-KR loss, dynamically adjust the weight $w$ and update the low-rank matrix $A$ and $B$ of LoRA. Combining mission-specific losses and mission-independent losses, the total gradient of LoRA parameters is:

$$\nabla_A \mathcal{L}_{\mathrm{LORA-KR}} = (1-w) \cdot \nabla_A \mathcal{L}_{\mathrm{CE}} + w \cdot \nabla_A \mathcal{L}_{\mathrm{KL}}$$

$$\nabla_B \mathcal{L}_{\mathrm{LORA-KR}} = (1-w) \cdot \nabla_B \mathcal{L}_{\mathrm{CE}} + w \cdot \nabla_B \mathcal{L}_{\mathrm{KL}}$$

5. Iterative optimization: Iterate the above process until the model converges.

The pseudocode for the algorithm is provided in the table 1 below.

Table 1. The pseudocode for the iterative optimization execution

| **Iterative optimization execution of the whole process** |
|---|
| **Input**: Query requirement Q and entity word E |
| 1: $\boldsymbol{B} = \boldsymbol{R}^{\boldsymbol{d \times r}}$ and $\boldsymbol{A} = \boldsymbol{R}^{\boldsymbol{r \times k}}$ |
| 2: **while** (1): |
| 3: Compute q and e according to $q$=BERT$_{\mathrm{Q}}$($Q$) and $e$=BERT$_{\mathrm{E}}$($E$) |
| 4: Compute $\hat{\boldsymbol{C}}$ according to $\hat{\boldsymbol{C}}$=Softmax($W_c$·z+A·(B·z)+$b_c$ ) |
| 5: Compute LORA-KR loss |
| 6: Compute $w$、$A$、$B$、$\boldsymbol{\nabla_A \mathcal{L}_{\mathrm{ProReg}}}$ and $\boldsymbol{\nabla_B \mathcal{L}_{\mathrm{ProReg}}}$ |
| 7: **if** （lim$\boldsymbol{\nabla_A \mathcal{L}_{LORA-KR}} \rightarrow \boldsymbol{0}$ and lim$\boldsymbol{\nabla_B \mathcal{L}_{LORA-KR}} \rightarrow \boldsymbol{0}$） **then:** |
| 8: **break;** |

By calculating the LORA-KR loss and the corresponding gradient and using a mechanism to dynamically adjust the weight $W$, we were able to fine-tune the model not only to accommodate the specific task domain, but also to maintain the model's memory of the original pre-trained domain knowledge, resulting in better generalization ability and task alignment. This method efficiently updates low-rank matrices $A$ and $B$ under the framework of LoRA, and finally improves the performance of the model in different fields.

## 4.3 Hierarchical Task-Based Agent

When fine-tuning LLMs, users may input unknown or highly complex requirements. Due to the LLM's limited knowledge and cognition, it may produce hallucinatory phenomena, such as confusing maintenance solutions for different parts, thereby providing incorrect answers. To overcome this challenge, this study adopts task-hierarchical Agent technology to enhance the LLM's capabilities by introducing external knowledge. In this architecture, the LLM acts as the "central brain," organizing and summarizing information, while the Agents serve as "assistants," providing additional, specific information support.

Specifically, the Agent's function can be represented by the following formula:

$$S_{t+1} = LLM(S_t, A_t, E_t)$$

where $S_t$ represents the state at time t。 $A_t$ is the action of the Agent， $E_t$ represents the state of the environment at time t， and $S_{t+1}$ is the new state at time t+1. The LLM is responsible for processing this information and updating the status.

The actions of the Agent can be represented by the following regularized function:

$$A_t = LLM(S_t, O_t)$$

where the LLM generates action $A_t$ based on the current state $S_t$ and the observation $O_t$.

Based on the Langchain framework, three agents with different functions were designed, all using the fine-tuned ChatGLM3 model as the base:

(1) Agent 1 is responsible for parsing user input related to maintenance tasks. If the user provides known maintenance objects, the task is assigned to Agent2, which has a database for these objects; For unknown maintenance objects, if the user inputs requirements for unknown maintenance objects, Agent1 analyzes the content. If it identifies similarities between the unknown and a known object, it associates them and sends the task to Agent2.

(2) Agent 2 provides customized solutions for various objects. When faced with different sub-tasks, Agent2 selects the appropriate tools for solving each task, ensuring that the solutions are professional and relevant.

(3) Agent 3 receives the solutions from Agent 2 and sends them back to the base LLM for further optimization. This ensures that the solutions are better adapted to the actual maintenance needs and scenarios.

Through this design, the LLM-R can accurately generate and optimize schemes for known maintenance objects. When encountering unknown objects or complex problems, it can analyze user needs through the collaborative work of the LLM and Agents, retrieve maintenance schemes for similar objects, and further customize these schemes using the advanced generation capabilities of the LLM.

# 5. Experiments

The experimental process is divided into three tasks: fine-tuning, maintenance scheme generation, and a question-and-answer case presentation. These tasks aim to evaluate the proposed method's effectiveness in generating maintenance suggestions and to enhance the model's accuracy and adaptability in this field.

(1) Fine-tuning experiments: This task evaluates the effectiveness of fine-tuning with LORA-KR loss and verifies the model's adaptability to new domains. Firstly, the LLM-F model of initial fine-tuning serves as the baseline, with its generation capabilities compared before and after fine-tuning. Secondly, the optimal balance point of the LLM-R under different data proportions is analyzed. Finally, the adaptability and accuracy of LLM-R and LLM-F in new fields are compared through controlled experiments.

(2) Maintenance scheme generation experiment: This task evaluates the generalization ability of the task-hierarchical agent combined with RAG for complex maintenance tasks and small sample objects. Two datasets are used: one with "known tasks" and another with "unknown task objects& similar known task objects." These datasets are used to conduct controlled experiments, verifying the model's ability to generate solutions with small samples. A multi-task mixed dataset is also created for robustness testing, ensuring performance stability across different domains.

(3) Question-and-answer maintenance proposal Case presentation: This task demonstrates the model's capability to generate maintenance recommendations in a question-and-answer format. It aims to provide task-related tips and guidance, comparing the model's performance to the traditional searchable IETM workflow. This comparison verifies the model's practicality and user experience benefits.

## 5.1 Construction and optimization of datasets

In this study, a comprehensive LLM dataset configuration strategy was adopted, designed to maintain the generality of the fine-tuned original model by mixing different proportions of self-built and public datasets.

(a) Maintenance scheme data set: This data set is the core of the fine-tuning process and contains approximately 1.8k entries. It includes maintenance schemes for five different objects: aircraft, trains, installers, agitators, and generators. Each entry is organized in a Q&A format. The cleaning and construction of these data provide a rich maintenance background and scenarios for the model.

(b) Alpaca dataset: The Alpaca[46] contains text data from various areas of life. It is divided into subsets of 4,000, 10,000, 14,000, and 20,000 entries for different fine-tuning needs. Introducing these data into the training process helps maintain the model's sensitivity to special domain knowledge while preserving its general knowledge capabilities.

## 5.2 Baselines

An experimental comparison was conducted between LLM-R and ChatGLM3-6B-based fine-tuned large models (LLM-F) to assess their ability to handle maintenance solution generation tasks in specific scenarios.

Table 2. Parameters and experimental tasks of LLM-R, LLM-F, Baichuan2, Baichuan2, LLaMa-2, Qwen2

| **Model** | **#params** | **Fine-tuning** | **Generate Maintenance Scheme** |
|---|---|---|---|
| LLM-R | 6B | √ | √ |
| LLM-F | 6B | √ | √ |
| Baichuan2[47] | 7B | × | √ |
| LLaMa-2[48] | 70B | × | √ |
| Qwen2[49] | 7B | × | √ |

Additionally, LLM-R and LLM-F were compared with several open-source LLMs to evaluate their security and language suitability for domain-specific data, particularly in small sample scenarios. The LLMs selected for this comparison include Baichuan2-7B, Llama2, and Qwen2-7B, as shown in Table 2.

## 5.3 Evaluation Metrics

In this study, three dimensions of evaluation metrics were adopted to comprehensively assess model performance. These metrics encompass traditional text generation evaluations, model-based evaluations, and human evaluations, covering a broad range of considerations. To be specific:

Traditional text generation evaluation metrics: ROUGE 1[50], ROUGE 2[50], ROUGE L[50] and BLEU[51] were used. These metrics evaluate the degree of overlap between the model-generated text and the reference text. ROUGE 1 focuses on the overlap of individual words, ROUGE 2 on the continuity between two words, and ROUGE L and BLEU on the similarity of longer word sequences and entire sentence structures.

Model-based evaluation indicators: BERT Score[52] and GPT[19] based on the BERT model were introduced to calculate the similarity between the generated text and the reference text. BERT Score includes Recall (R), Precision (P), and F1 Score (F), measuring both text and semantic similarity to assess the quality and accuracy of the content. GPT scores evaluate the naturalness and fluency of the generated text using a pre-trained language model.

Human-based evaluations: For a more comprehensive assessment of model performance, human evaluations were included. From the perspective of the experience and cognition of maintenance personnel, this evaluation method examines whether the text generated by the model conforms to the maintenance background, whether it is smooth and natural, and whether it is accurate.

These comprehensive evaluation metrics not only assess the model's capabilities from different perspectives but also evaluate the generated text at both micro and macro levels.

## 5.4 Experimental Settings

Table 3 and Table 4 show all the key parameter Settings used in this experiment. These parameters cover computer configuration, LLM basic parameters, fine tuning parameters, and specific parameter selection for RAG technology. Below are the parameters that form the basis of the experiment:

Fine-tuning parameters: Fine-tuning is the process of adjusting a pre-trained model to handle specific maintenance tasks. At this stage, parameters such as the learning rate, LORA rank, epoch number, and batch size are refined to ensure the model effectively adapts to the new dataset.

Temperature coefficient setting: The temperature coefficient is an important parameter to control the diversity of text generated by the model. Higher temperature coefficients result in more diverse text, while lower coefficients produce more focused and consistent text.

RAG parameters: RAG technology combines two aspects, Retrieval and Generation, to enhance the ability of models to deal with complex problems. In this study, the parameters chunk_size=300 and chunk_overlap=50 was used.

Table 3. Fine-tuning parameters

| **Parameter1** | | | | | |
|---|---|---|---|---|---|
| max_length | learning rate | epoch | batch size | lora rank | temperature |
| 1024 | 1e-4 | 4 | 4 | 8 | 0.7 |

Table 4. System parameters

| **Parameter2** | | | | |
|---|---|---|---|---|
| Graphics | RAM | Development tool | Framework | Cuda version |
| RTX 4090 | 24GB | Pycharm | Torch 2.1.0 | 12.2 |

# 6 Experimental Results and Case Studies

## 6.1 Fine tuning experiment

In the specific field of maintenance, the use of LLMs requires precise fine-tuning to ensure they can meet the specific needs and knowledge of users in the field. The key to effective fine-tuning is utilizing special domain datasets, which make the model more accurate and relevant when dealing with maintenance-related expertise.

### 6.1.1 Fine tuning of Special field

First, the dataset described in Section 4.1 was used, with a ratio of special domain data to general domain data set at 1:0. The LLM was fine-tuned using this proportional dataset. The results of the fine-tuning are shown in Table 5, and the model loss rate during the fine-tuning process is illustrated in Figure 4.

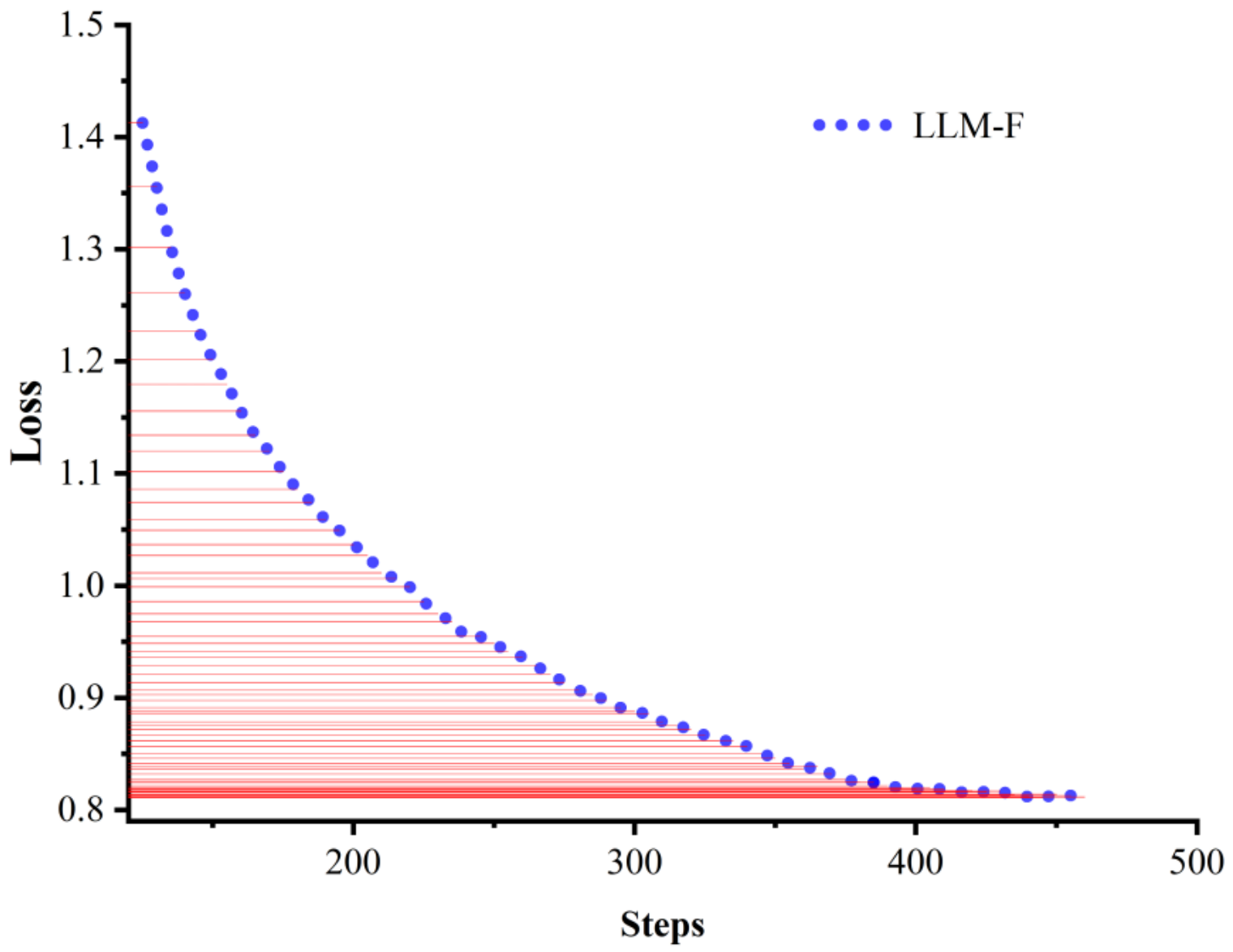

Figure 4. Loss ratios.

Table 5. Individual indicator scores

| | | |
|---|---|---|
| BERT score | Call ratio | 87.50% |
| | Accuracy | 87.81% |
| | F1 score | 87.64% |
| Rouge score | Rouge-1 | 71.53% |
| | Rouge-2 | 51.63% |
| | Rouge-L | 65.12% |
| BLEU | — | 60.27% |

In the above experiments, the self-built special domain dataset was used to test the model. The experimental results show that the fine-tuned LLM achieved significant improvements in BERT score, ROUGE score, and BLEU, indicating that using special domain data exclusively for fine-tuning is beneficial.

### 6.1.2 Fine-tuning of model capabilities

Fine-tuning of the LLM for a special domain introduces the challenge of catastrophic forgetting, where the model loses some of its original capabilities as it adapts to new domain-specific knowledge. To effectively solve this problem, a data sample mixing strategy was employed, as described in Sections 4.1 and 5.1. This strategy involves mixing domain-specific knowledge with general knowledge datasets in various proportions: 1:1, 1:2, 1:5, 1:7, and 1:10. The goal is to find a balance that allows the model to effectively absorb and adapt to new domain knowledge while retaining its original capabilities. With this strategy, the model can learn the special domain knowledge while preserving the applicability of its original capabilities.

Two sets of experiments were conducted. In the first experiment, training sets with different proportions were used to fine-tune the LLM, which was then tested using the self-built datasets. The results of these experiments are shown in Table 6.

Table 6. Individual indicator scores

| **Data** | **Rouge-1** | **Rouge-2** | **Rouge-L** | **Call ratio** | **Accuracy** | **F1 score** | **BLEU** |
|---|---|---|---|---|---|---|---|
| 1_0 | 71.53% | 51.63% | 65.12% | 87.50% | 87.81% | 87.64% | 60.27% |
| 1_1 | 71.05% | 51.23% | 64.87% | 87.77% | 87.51% | 87.63% | 60.95% |
| 1_2 | 73.11% | 54.24% | 68.37% | 87.88% | 87.90% | 87.88% | 60.58% |
| 1_5 | 72.69% | 53.73% | 65.93% | 87.38% | 87.65% | 87.50% | 61.02% |
| 1_7 | 73.82% | 54.87% | 67.47% | 87.82% | 87.91% | 87.85% | 63.43% |
| **1_10** | **74.24%** | **56.31%** | **69.14%** | **88.28%** | **89.03%** | **88.63%** | **64.50%** |

In the second experiment, we used common data to test the model, and the scores under each index are shown in Table 7.

Table 7. Individual indicator scores

| Data | Rouge-1 | Rouge-2 | Rouge-L | Call ratio | Accuracy | F1 score | BLEU |
|---|---|---|---|---|---|---|---|
| 1_1 | 33.5% | 10.41% | 27.17% | 69.53% | 73.80% | 71.55% | 23.08% |
| 1_2 | 32.1% | 8.29% | 23.53% | 68.26% | 73.56% | 70.77% | 18.36% |
| 1_5 | 32.2% | 8.60% | 26.49% | 69.10% | 71.63% | 70.29% | 22.36% |
| **1_7** | **34.3%** | **12.05%** | **24.51%** | **71.08%** | **75.17%** | **73.03%** | **27.89%** |
| 1_10 | 34.2% | 11.14% | 25.33% | 69.72% | 72.75% | 71.18% | 25.88% |

According to the experimental results in Table 6 and Table 7, the model scored highest at a 1:10 ratio when tested on the self-built dataset. However, it scored highest at a 1:7 ratio when tested on the general dataset. This indicates that the model performs differently on different test sets and at different scaling ratios. Figure 5 illustrates the performance of different test sets across various evaluation scores.

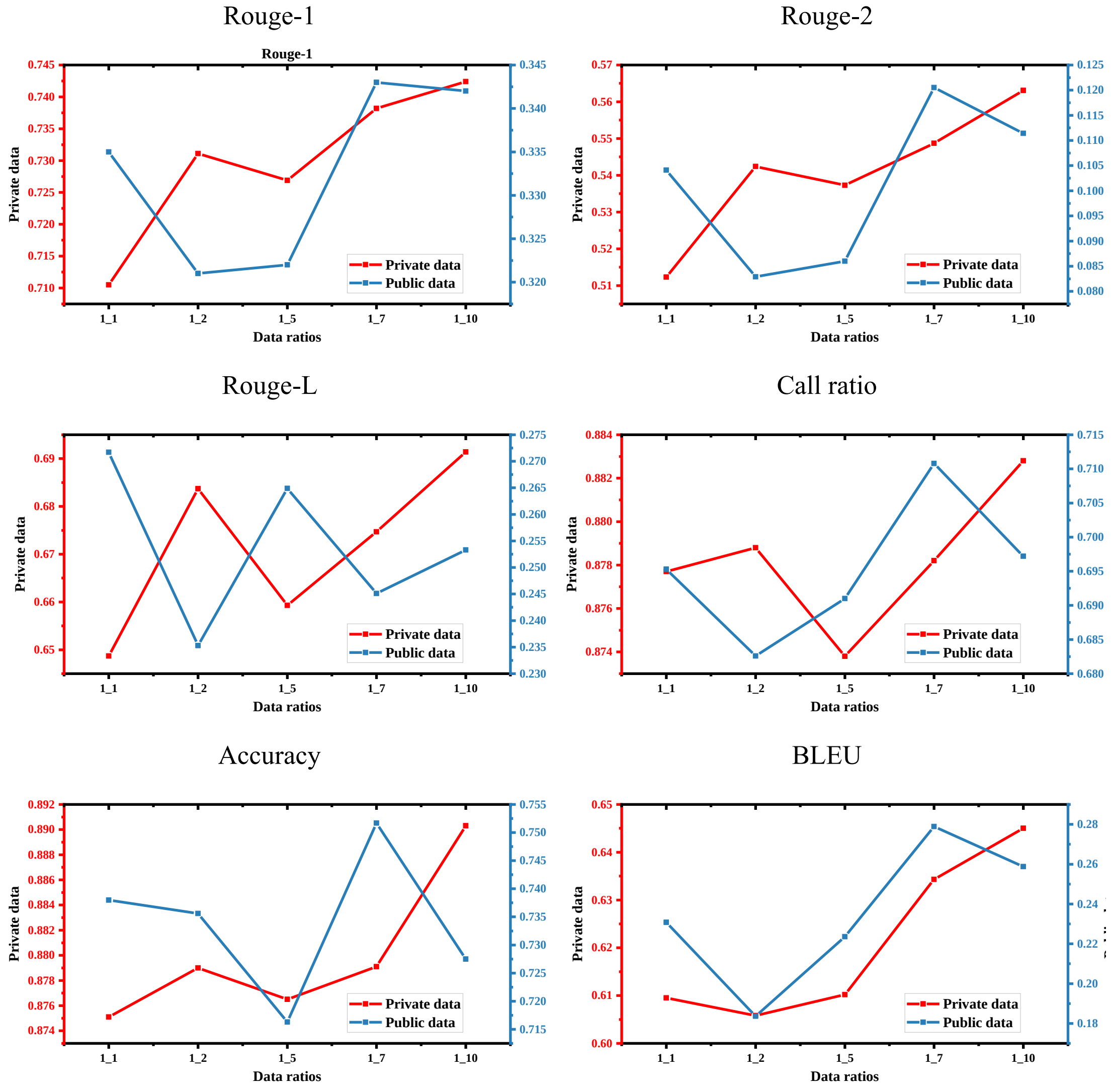

Figure 5. Individual indicator scores

To sum up, although the 1:10 scale model showed the best performance on the self-built dataset, the 1:7 scale model excelled on the general dataset. Therefore, considering factors such as training time, cost, data volume, and overall effectiveness, the 1:7 scale model was selected for subsequent experiments.

## 6.2 Experiment on Maintenance Scheme Generation

### 6.2.1 Comparative experiment

In this experiment, varying proportions of data were injected into the vector library in LLM-R, matching the data proportions used in the fine-tuning model (LLM-F) from Sections 4.1 and 5.1. The performance of the fine-tuning model LLM-F and the LLM-R was compared through experiments, with the results shown in Figure 6.

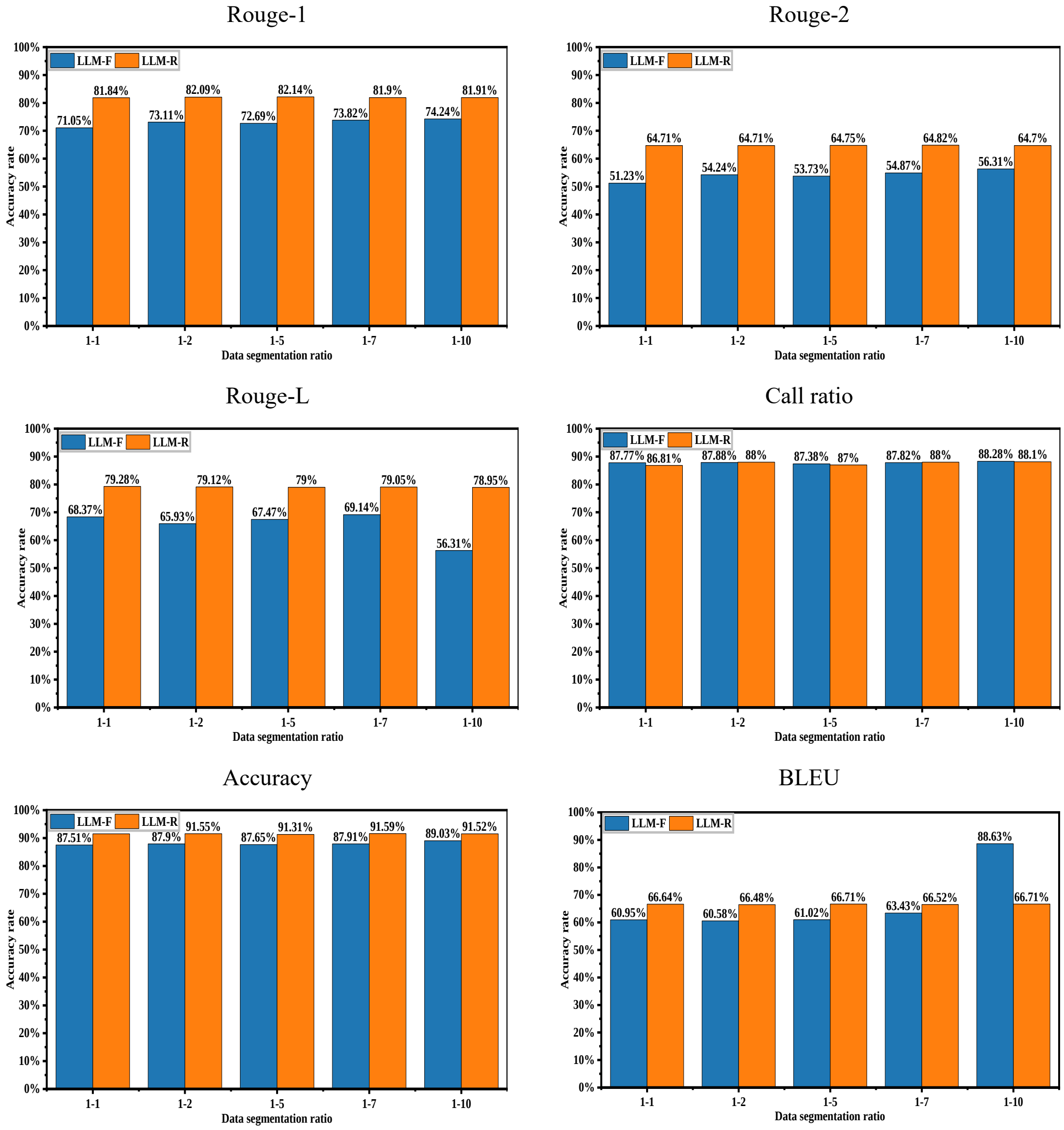

Figure 6. Individual indicator scores

The maintenance scheme results indicate that the LLM-R significantly outperforms the LLM-F model with full parameters, demonstrating higher stability when processing the same scale data. LLM-R uses Instruction-level RAG, a strategy that combines information retrieval and content generation capabilities. In this framework, LLM-R first retrieves information fragments closely related to the query from a vast knowledge base and then generates answers based on this information. This approach effectively avoids the "hallucination" phenomenon often observed in traditional LLMs (such as LLM-F), where generated information may not align with the facts. By directly using relevant pieces of information, LLM-R provides more accurate and reliable outputs. In addition, LLM-R introduces Hierarchical Task-Based Agent technology, an advanced task management mechanism that decomposes and organizes complex tasks more effectively and adjusts the processing strategy according to different stages. This hierarchical approach greatly enhances the accuracy and efficiency of the model in executing multi-step or expert-intensive tasks. Compared to LLM-F, LLM-R's incorporation of Instruction-level RAG technology prevents information misdirection, and the use of Hierarchical Task-Based Agents optimizes task execution, significantly improving the model's overall performance and stability.

### 6.2.2 Small samples experiment

In the actual maintenance process, maintenance personnel often encounter unknown objects. If these unknown objects are similar to known objects or if their maintenance schemes are similar, the known schemes can serve as a basis for providing maintenance solutions for the unknown objects. To verify the model's capability with small samples, manual and GPT evaluations were conducted.

Specifically, four state-of-the-art (SOTA) baselines were selected for comparison with LLM-R. Five controlled experiments were randomly conducted, each with training and test sets constructed from the datasets described in Sections 4.1 and 5.1. A small sample test set was used: "unknown task, unknown task object & similar known task object," with specific cases detailed in Section 6.3. This approach aimed to evaluate LLM-R's generalization ability under small sample conditions. The model was tested on unknown maintenance tasks by referencing maintenance schemes for known objects with similar structures or functions. The experimental results are summarized in Tables 8 and 9.

Table 8 Manual evaluation score

| Model | Score 1 | Score 2 | Score 3 | Score 4 | Score 5 | AVG |
|---|---|---|---|---|---|---|
| LLM-F | 6 | 5.5 | 5 | 3 | 4 | 4.75 |
| LLM-R | **7.5** | **10** | **8** | **7** | **8** | **7.5** |
| Baichuan | 2 | 2.5 | 1.7 | 2 | 1.5 | 1.77 |
| Llama | 1.5 | 1 | 2 | 1.1 | 1.7 | 1.46 |
| Qwen | 2.5 | 1 | 2 | 2.5 | 1.5 | 1.75 |

The experimental results demonstrate that the proposed LLM-R achieved the highest scores in both manual and GPT evaluations. Notably, the accuracy rate improved by 23% over the baseline ChatGLM3, outperforming other LLMs. These results fully validate that LLM-R can effectively generate preliminary maintenance schemes.

Table 9 GPT evaluation score

| Model | Score 1 | Score 2 | Score 3 | Score 4 | Score 5 | AVG |
|---|---|---|---|---|---|---|
| LLM-F | 8 | 7.5 | 8 | 7.5 | 8 | 7.75 |
| LLM-R | **9** | **8.5** | **9.5** | **9** | **8** | **8.8** |
| Baichuan | 6.5 | 6 | 6 | 5.5 | 6.5 | 5.6 |
| Llama | 4 | 3 | 3.5 | 4 | 5 | 4.1 |
| Qwen | 4.5 | 5 | 4.5 | 4 | 6 | 4.95 |

### 6.2.3 Experiments in different fields

The purpose of this experiment is the performance of LLM-R in generative question answering and accurate information retrieval within a simulated complex maintenance environment. It also seeks to verify LLM-R's efficiency and accuracy improvements over traditional IETM search methods. To evaluate the model's cross-domain applicability, a comprehensive maintenance task dataset was constructed from the datasets in Sections 4.1 and 5.1, covering multiple objects (such as aircraft, trains, and agitators). The experiment involved maintenance schemes for more than five different types of objects, aiming to demonstrate that LLM-R can handle not only single-type maintenance tasks but also adapt to and effectively solve the maintenance problems of diverse objects. The experimental results (shown in Figure 7) indicate that LLM-R achieves the highest accuracy across all types of objects, preliminarily proving its wide adaptability and high efficiency in maintenance tasks.

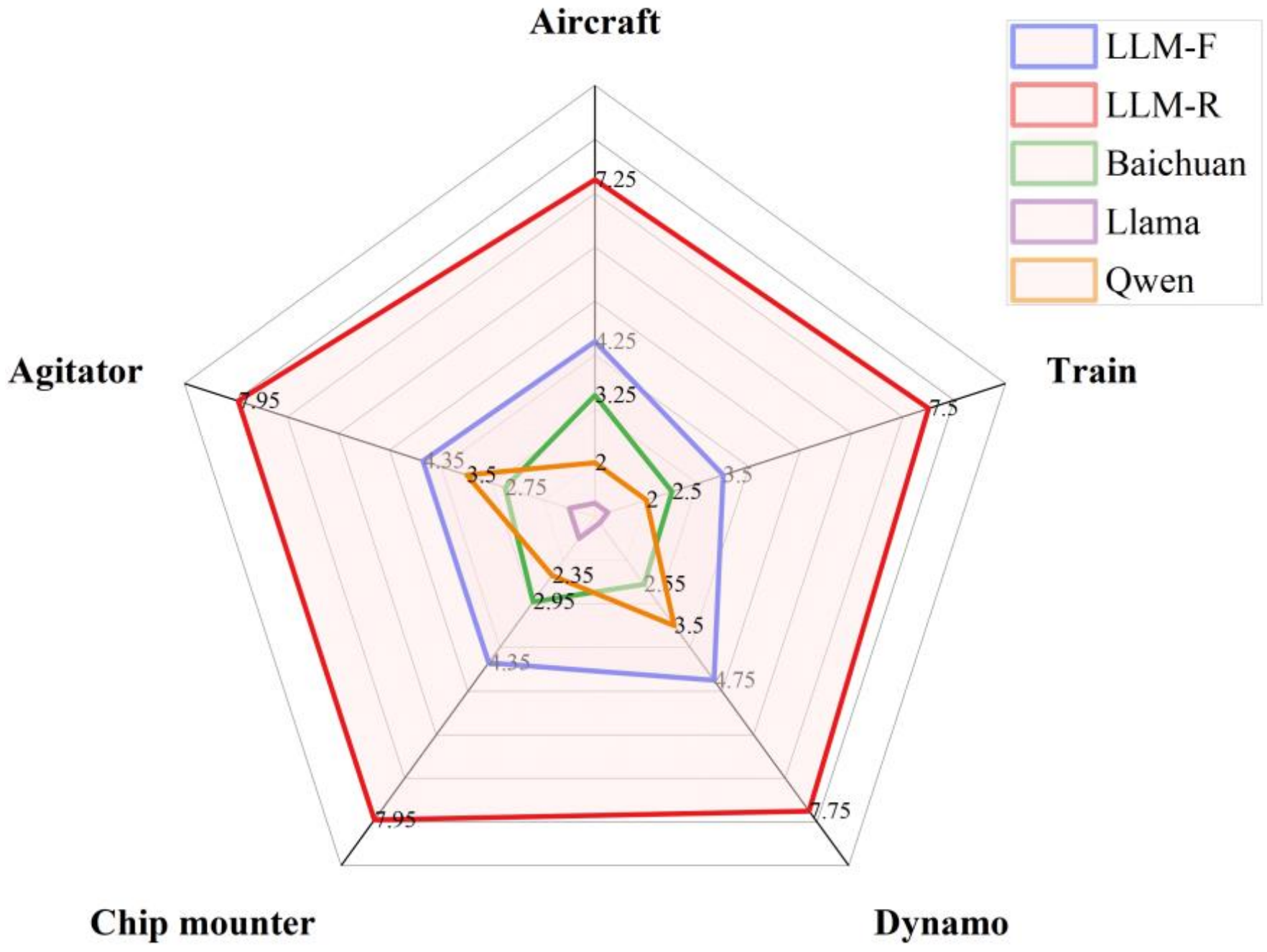

Figure 7. The representation of the model on different objects

## 6.3 Case Studies

This section highlights the differences between IETM and LLM-R in information acquisition,

scheme screening, and further search when dealing with normal and small sample maintenance problems. The advantages of the LLM-R system in information analysis and providing detailed operation steps are evident, as shown in Table 10 for IETM and normal sample cases, and Table 11 for small sample cases.

Table 10. Case comparison of normal samples

| **Interaction Phase** | **IETM** | **LLM-R** |
|---|---|---|
| Problem Description | User opens the IETM system and inputs the keyword “engine starting difficulty. ” | User opens the LLM-R system and describes the problem: “The engine has difficulty starting in cold weather and there is a noise.” |
| Information Retrieval | System returns 50 related documents. User browses the document titles one by one. | The system analyzes the information. |
| Information Screening | User reads the documents and extracts part of the Maintenance Scheme, including checking the fuel pump, ignition system, ECU, and fuel pressure. However, the scheme is incomplete and requires further searching. | 1. Check if the fuel pump and fuel filter are clogged and replace them if necessary.<br>2. Verify that the ignition system is working properly; inspect the spark plugs and ignition coils.<br>3. Use diagnostic tools to read the fault codes from the Engine Control Unit (ECU).<br>4. Test the fuel pressure at low temperatures and replace the fuel pressure regulator if there are abnormalities.<br>5. Check the engine oil viscosity to ensure it is suitable for low-temperature environments. |
| Further Search | User inputs a new keyword “cold start. ” System returns 30 documents. User screens and reads the documents again. | User encounters specific problems during the maintenance process and asks the system questions. System immediately provides specific operational steps and illustrations. |

Table 11. Small sample case

| Interaction Phase | LLM-R |
|---|---|
| Problem Description | User opens the LLM-R system and describes the problem: "The new hydraulic pump has a failure. The similar and known maintenance scheme is for the fuel pump." |
| Information Retrieval | The system analyzes the information. |
| Information Screening (Maintenance Scheme Comparison) | 1. Check if the power connection of the hydraulic pump is normal.<br>2. Verify if the hydraulic pump control unit is working properly.<br>3. Check and clean the hydraulic pump filter.<br>4. For insufficient pressure, recommend replacing the pressure regulator inside the hydraulic pump. |

## 7. Conclusions

Through a detailed analysis of the current maintenance field needs and LLM-related technologies, this study has successfully developed an automatic maintenance scheme generation system based on LLMs. The proposed LLM-R integrates a Supervised Fine-Tuning of LORA-KR loss technology, Hierarchical Task-Based Agent and Instruction-level RAG technology. This combination enables the accurate decomposition of maintenance tasks and the generation of corresponding solutions while significantly reducing hallucinations. The model was tested on datasets from various fields, achieving an average accuracy rate of 91.59%, demonstrating its ability to understand complex linguistic logic in maintenance data and generate accurate maintenance schemes.

Despite these achievements, several challenges remain for broader application. The accuracy of maintenance scheme outputs and vectorized search capabilities of LLM-R need further enhancement. Additionally, the lack of contextual memory in LLM-R affects the efficiency of generating maintenance schemes in multi-round dialogues. Future research could focus on:

Improvement of the output accuracy of the maintenance scheme: By introducing a more comprehensive professional maintenance knowledge base and practical maintenance cases, the output accuracy of LLM-R can be improved. Implementing Muti-expert model technology will allow for the dynamic selection of the most suitable sub-model according to specific maintenance tasks, thereby enhancing the relevance and effectiveness of the generated schemes.

Optimization of vectorized search algorithm: To improve the accuracy and speed of retrieving maintenance schemes, more advanced vectorized search algorithms, such as self-supervised learning and comparative learning, should be adopted. Combining these algorithms with graph neural networks (GNN) can help to delve deeper into historical maintenance data, improving the relevance and accuracy of search results.

Enhancement of man-machine collaborative memory ability: By leveraging the memory technology of the Agent, LLM-R's ability to remember context across multiple rounds of dialogue can be enhanced. This will allow the system to maintain continuity from previous interactions, thereby improving the efficiency and coherence of generating maintenance schemes.

The method can be widely used in aviation, manufacturing, energy, transportation, and other fields to enhance equipment operation efficiency and safety. It can reduce downtime, lower maintenance costs, and extend equipment lifespan, thereby improving the productivity and market competitiveness of enterprises and significantly boosting economic efficiency.

# 8 Declaration

**Competing Interests:** The authors have no relevant financial or non-financial interests to disclose.

**Author Contributions:** Data curation, Q.H., B.L. C.L. and X.H.; Formal analysis, L.T.; Methodology, Q.H. and L.T.; Writing—original draft, Q.H., W.Z., B.L. and Y.W.; Validation, B.L.; Investigation, B.L. and Q.H.; Supervision, L.T., X.W. C.L. and X.H. All authors have read and agreed to the published version of the manuscript.